\documentclass[runningheads]{llncs}
\usepackage[T1]{fontenc}
\usepackage{graphicx}
\begin{document}
\title{Cyber Humanism in Education: Reclaiming Agency through AI and Learning Sciences}
\titlerunning{Cyber Humanism in Education}
%
\author{Giovanni Adorni\orcidID{0000-0003-3933-4377}}
\authorrunning{G.Adorni}
%
\institute{DIBRIS, University of Genoa, \\ Viale F. Causa 13, 16145 Genoa - Italy\\
\email{giovanni.adorni@unige.it}}
\maketitle              
\begin{abstract}
Generative Artificial Intelligence (GenAI) is rapidly reshaping how knowledge is produced and validated in education. Rather than adding another digital tool, large language models reconfigure reading, writing, and coding into hybrid human-AI workflows, raising concerns about epistemic automation, cognitive offloading, and the de-professiona\-lisation of teachers. This paper proposes \emph{Cyber Humanism in Education} as a framework for reclaiming human agency in this landscape. We conceptualise AI-enabled learning environments as socio-technical infrastructures co-authored by humans and machines, and position educators and learners as epistemic agents and \emph{algorithmic citizens} who have both the right and the responsibility to shape these infrastructures.

We articulate three pillars for cyber-humanist design, \emph{reflexive competence}, \emph{algorithmic citizenship}, and \emph{dialogic design}, and relate them to major international digital and AI competence frameworks. 
We then present higher-education case studies that operationalise these ideas through \emph{prompt-based learning} and a new \emph{Conversational AI Educator} certification within the EPICT ecosystem. The findings show how such practices can strengthen epistemic agency while surfacing tensions around workload, equity, and governance, and outline implications for the future of AI-rich, human-centred education.
\end{abstract}
\section{Introduction}
\label{sec:introduction}

%
Generative Artificial Intelligence (GenAI) is rapidly becoming embedded in the everyday practices of students and educators. Large language models and multimodal systems are integrated in search engines, productivity tools, and learning platforms, and are used for writing, coding, summarising, and ideation. From a humanistic and Learning Sciences perspective, GenAI does not simply add another ``tool'' to the digital repertoire of schools and universities: it reconfigures what counts as cognitive work, shifting segments of reading, writing, translation, and coding into hybrid human--AI workflows. Recent work in Cyber Humanism argues that intelligent infrastructures should be seen as co-authors of culture and knowledge, rather than neutral channels or external instruments \cite{CHManifesto}. Education is therefore confronted with an epistemic and infrastructural challenge: how to preserve and expand human agency when core activities are increasingly mediated by opaque, data-driven systems.

Research and policy reports highlight both risks and opportunities. At the cognitive level, the automation of reasoning and text production may lead to \emph{epistemic automation} and excessive \emph{cognitive offloading}, where learners outsource problem solving and judgement to systems they only partially understand. At the professional level, there is concern about the de-professionalisation of teachers, whose expertise risks being reduced to supervising AI-generated content while planning, feedback, and assessment are delegated to algorithms; occupational health analyses point to additional challenges related to workload, digital stress, surveillance, and erosion of autonomy \cite{EUOSHAAI,UNESCOAIHE}. At the same time, AI can scaffold metacognition, support personalised feedback, and enable new forms of collaborative inquiry if used critically and transparently. The key question is therefore not whether AI should be used in education, but how it can be integrated in ways that strengthen, rather than weaken, the epistemic agency and civic responsibility of students and educators.

This paper proposes \emph{Cyber Humanism in Education} as a framework to address this tension. Cyber Humanism reframes AI-enabled learning environments as socio-technical infrastructures co-authored by humans and machines. Instead of treating educators and learners as mere users or consumers of intelligent systems, it positions them as epistemic agents and \emph{algorithmic citizens} who have the right and responsibility to understand, interrogate, and shape the AI systems that mediate their learning.

Building on this perspective, the paper addresses three guiding \textbf{research questions}:
\vspace{-8pt}
\begin{itemize}
  \item \textbf{RQ1:} How can Cyber Humanism reframe the role of AI in education from a disruptive external force to a set of co-designed learning infrastructures?
  \item \textbf{RQ2:} Which competencies and frameworks are needed to support educators and learners as epistemic and algorithmic citizens in AI-rich learning environments?
  \item \textbf{RQ3:} How can prompt-based learning and conversational AI certification instantiate these principles in higher education practice?
\end{itemize}

\noindent
To address these questions, the paper makes three contributions. First, it develops a conceptual lens that connects Cyber Humanism with core constructs from the Learning Sciences, such as epistemic agency, metacognition, and dialogic learning, and articulates three pillars for reclaiming agency in AI-enabled education: \emph{reflexive competence}, \emph{algorithmic citizenship}, and \emph{dialogic design}. Second, it provides a concise mapping of major AI- and digital-competence frameworks, including DigComp 3.0 \cite{DigComp30}, DigCompEdu AI Pioneers \cite{AIPioneers}, UNESCO's AI competency frameworks \cite{UNESCOteachers,UNESCOstudents}, and the OECD/EC AI Literacy framework \cite{AILitFramework}, identifying gaps regarding epistemic agency, infrastructural participation, and the role of natural language as a problem-solving and computational medium. Third, it illustrates how Cyber Humanism can be operationalised through case studies in higher education, focusing on \emph{prompt-based learning} (PBL), where natural language interactions with GenAI are used as a bridge between problem description, problem solving, and computational thinking, and on the design and early implementation of a \emph{Conversational AI Educator} certification within the EPICT ecosystem.

The remainder of the paper is organised as follows. Section~\ref{sec:cyberhumanism} traces the evolution from Digital Humanism to Cyber Humanism and introduces the key constructs that underpin our educational perspective. Section~\ref{sec:frameworks} reviews the current policy and competency landscape for AI in education, focusing on European and global frameworks and identifying gaps from a cyber-humanist standpoint. Section~\ref{sec:designlens} presents Cyber Humanism as a design lens for AI-enhanced learning, articulating the three pillars and introducing prompt-based learning as a bridge between natural language and computational thinking. Section~\ref{sec:casestudies} reports case studies from higher education that operationalise these ideas through prompt-based learning and the EPICT Conversational AI Educator certification \cite{EPICT}. Section~\ref{sec:discussion} discusses how these cases contribute to reclaiming agency in AI-rich educational ecosystems, highlighting tensions and implications for the Learning Sciences. Finally, Section~\ref{sec:conclusion} concludes the paper by summarising the main contributions and outlining directions for future research and practice on Cyber Humanism in Education.

\section{Cyber Humanism in Education}
\label{sec:cyberhumanism}
\subsection{From Digital Humanities to Digital Humanism}

The relationship between computing and the humanities has evolved from early \emph{Digital Humanities} (DH), where computers were mainly used to support existing scholarly practices (e.g., digital editions, text encoding, corpus analysis), to broader engagements with digital culture, platforms, and datafication. In this second phase, the digital environment itself became an object of critical inquiry, and humanists turned their attention to issues such as platform power, participatory culture, and algorithmic mediation. Yet computational systems were still often treated as relatively stable environments within which cultural practices unfolded, rather than as dynamic, co-evolving agents in the production of meaning.

In response to the growing influence of large-scale platforms and algorithmic systems, the notion of \emph{Digital Humanism} emerged, articulated for example in the Vienna Manifesto \cite{ViennaManifesto}. Digital Humanism emphasises the need to place human values, rights, and democratic principles at the centre of digital transformation, calling for regulatory frameworks, ethical guidelines, and socio-technical arrangements that ensure technologies serve the public good. This move has been crucial in foregrounding questions of justice, inclusion, and democracy in debates on AI and digitalisation.

However, Digital Humanism tends to preserve a conceptual separation between humans and technologies, casting the former primarily as subjects to be protected and the latter as objects to be regulated. Technologies appear as external forces that must be tamed or aligned with human values. In educational debates, this often translates into a defensive focus on preventing harm (e.g., plagiarism, bias, surveillance), while leaving largely unexamined how AI systems reshape the infrastructures, practices, and epistemic norms of teaching and learning.

\subsection{Cyber Humanism, Algorithmic Citizenship, Epistemic Agency}

\emph{Cyber Humanism} has been proposed as a further step in this trajectory, one that takes seriously the entanglement of humans and computational systems in the co-production of knowledge, culture, and institutions \cite{CHManifesto}. Rather than framing the digital as an external environment, Cyber Humanism understands digital infrastructures, AI models, and data ecosystems as \emph{cognitive infrastructures}: they participate in shaping how problems are formulated, which solutions appear plausible, and whose epistemic contributions are recognised. In this view, AI systems are not neutral tools but situated agents in socio-technical networks whose behaviour can stabilise or destabilise existing power relations and epistemic hierarchies.

A key concept within Cyber Humanism is that of \emph{algorithmic citizenship}. While traditional notions of digital citizenship emphasise access, participation, and responsibility in online spaces, algorithmic citizenship foregrounds the rights and responsibilities of individuals and communities vis-à-vis the algorithmic infrastructures that shape their opportunities and obligations. Algorithmic citizens are not only competent users of digital services; they are also stakeholders who can interrogate, critique, and participate in the design and governance of the systems that affect their lives. Translating this idea into education entails reimagining learners and educators not as passive recipients of AI-enhanced tools, but as epistemic agents with a legitimate voice in how AI is deployed in curricula, assessment, and institutional governance.

This perspective resonates with core concerns in the Learning Sciences. Research on epistemic cognition and epistemic agency has long emphasised that learners are not merely consumers of knowledge but active participants in its construction, evaluation, and communication \cite{KnowledgeBuilding}. Epistemic agency refers to the capacity of individuals and communities to initiate and regulate knowledge-building processes: deciding which questions are worth pursuing, what counts as evidence, and how to justify claims within a community of practice. Dialogic and socio-cultural approaches to learning further stress that knowledge emerges through interaction and negotiation in specific cultural and material contexts \cite{Wells99,Sfard19}. Tools and artefacts, including digital ones, play a central role in mediating these interactions and in shaping the space of possible actions and interpretations.

When AI systems enter this ecology, they do not merely add new resources; they reconfigure the conditions under which epistemic agency is exercised. Learners who rely on generative models to formulate arguments, summarise readings, or generate code engage in a form of co-authorship with systems whose internal workings are opaque and whose training data are largely inaccessible. From a cyber-humanist standpoint, the key educational challenge is to design AI-rich learning environments that expand rather than erode epistemic agency. This requires making visible how AI systems participate in knowledge construction, equipping learners and educators with the competences needed to critically interrogate and reshape these mediations, and integrating questions of infrastructure and governance into the Learning Sciences agenda.

In this paper, we take Cyber Humanism as a bridge between these debates. By framing AI as part of the cognitive and institutional infrastructures of education, Cyber Humanism invites the Learning Sciences to engage more directly with algorithmic citizenship and infrastructural design. Conversely, the Learning Sciences provide Cyber Humanism with fine-grained concepts and empirical methods to understand how epistemic agency is negotiated in everyday human--AI interactions. The following sections build on this convergence to analyse existing AI and digital competence frameworks, and to articulate design principles and practices---such as prompt-based learning and conversational AI certification---that aim to reclaim agency in AI-enabled education.

\section{Policy and Competency Landscape for AI in Education}
\label{sec:frameworks}
As generative AI becomes embedded in educational practice, policy-makers and international organisations have responded by developing digital and AI competence frameworks for citizens, students, and educators. These frameworks are not neutral catalogues of skills: they function as normative infrastructures that define what it means to be a competent digital subject and, increasingly, an AI-literate learner or teacher. From a cyber-humanist perspective, they reveal how institutions imagine the relationship between humans and intelligent systems, and which forms of agency and responsibility are foregrounded or marginalised.

\subsection{European Digital and AI Competence Frameworks}

The European Commission has invested in a family of competence frameworks for a digitally transformed society. \emph{DigComp} provides a general framework for digital competence for all citizens, while \emph{DigCompEdu} \cite{DigCompEdu} focuses on educators' professional practice. Recent versions explicitly integrate AI-related competence.

DigComp 3.0 conceptualises digital competence as a set of capabilities organised in five areas, covering information and data literacy, communication and collaboration, digital content creation, safety and wellbeing, and problem solving and continuous learning \cite{DigComp30}. AI is treated as a cross-cutting concern, implicitly present in how information is accessed and curated and in how digital systems support collaboration, creativity, and decision-making. Area~5 is particularly relevant here, as it emphasises problem identification and solving, innovation, and the capacity to learn and adapt in evolving digital environments, explicitly mentioning computational thinking and the ability to work with automated systems.

DigCompEdu complements this citizen-oriented perspective with a framework targeted at educators. It defines competence areas ranging from professional engagement and digital resources to teaching, assessment, and empowering learners \cite{DigCompEdu}. Recent initiatives extend DigCompEdu with AI-focused descriptors, highlighting the need for educators to understand opportunities and risks of AI, to design learning activities that integrate AI responsibly, and to model critical and ethical engagement with intelligent systems. Projects such as AI Pioneers build on this foundation to develop instruments for assessing institutional readiness, governance, and pedagogical use of AI, linking individual competence with organisational strategies. Overall, these frameworks articulate a vision of the AI-competent educator as a reflective professional who can select, adapt, and orchestrate technologies in pedagogically meaningful ways, but they say relatively little about active participation in the design and governance of AI infrastructures.

\subsection{Global AI Literacy and Competency Frameworks}

International organisations have proposed more explicitly AI-focused competency models. UNESCO has developed AI competency frameworks for both students and teachers. The student framework articulates AI literacy in terms of knowledge, skills, and values, emphasising understanding of basic AI concepts, applications, and societal implications, as well as ethical awareness and a commitment to human rights and sustainable development \cite{UNESCOstudents}. The teacher framework defines dimensions such as a human-centred mindset towards AI, the ethics of AI in education, foundational AI knowledge, AI-enhanced pedagogy, and AI for teachers' own professional learning \cite{UNESCOteachers}. Both frameworks position AI as a public good and stress equity, inclusion, and cultural diversity.

The OECD and the European Commission have advanced an \emph{AI Literacy} (AILit) framework targeting a broad public, including learners and educators \cite{AILitFramework}. AILit typically distinguishes domains such as engaging with AI systems, using AI productively, managing AI-related risks and impacts, and, at more advanced levels, participating in the design and shaping of AI systems. This moves beyond purely consumer-oriented notions of literacy by acknowledging that people may play different roles in AI ecosystems, from end-users to co-creators and stewards. These global frameworks converge in portraying AI literacy as multi-dimensional, combining conceptual understanding, practical skills, ethical dispositions, and civic awareness. However, they offer limited guidance on how AI-related competencies can be enacted in concrete classroom practices, especially in higher education.

\subsection{Higher Education Challenges and Cyber-Humanist Gaps}

Higher education institutions face specific challenges in integrating AI in ways that align with their missions of teaching, research, and public engagement. Recent analyses underscore that universities are simultaneously sites of AI innovation, early adopters of AI-enhanced tools, and custodians of academic standards and integrity. UNESCO reports point to a persistent AI skills gap among both students and staff, inconsistent institutional policies, and uncertainties around the legal and ethical status of AI-generated content \cite{UNESCOAIHE}. At the same time, there is pressure to leverage AI for efficiency, personalisation, and competitive advantage, sometimes driven more by vendor narratives than by evidence from the Learning Sciences. Occupational health and safety perspectives add another layer, highlighting risks related to workload intensification, continuous monitoring, loss of autonomy, and new forms of digital stress and techno-anxiety for educators \cite{EUOSHAAI}.

From a cyber-humanist standpoint, important gaps remain across these frameworks and reports. Most still conceptualise AI primarily as a set of tools or services to be used responsibly, rather than as part of the cognitive and institutional infrastructures that shape how knowledge is produced, validated, and distributed. Learners and educators are cast as informed users and, at best, co-creators within predefined platforms, but rarely as algorithmic citizens with a say in how AI systems are designed, deployed, and governed in their educational environments. Moreover, the micro-level practices through which AI literacy and competence are enacted are only loosely specified. Few frameworks engage with the role of natural language as a primary medium for interacting with generative models, and with the opportunities this creates for rethinking problem formulation, computational thinking, and collaborative inquiry. Nor do they systematically address how professional roles, such as that of a ``Conversational AI Educator'', could be institutionalised and supported through certification and organisational change. These gaps motivate the cyber-humanist approach developed in the rest of this paper.

\section{Cyber Humanism: Design Lens for AI-Enhanced Learning}
\label{sec:designlens}
The previous sections have shown that current digital and AI competence frameworks provide important reference points for policy and curriculum design, but leave partly open how epistemic and algorithmic agency can be cultivated in concrete educational practices. In this section, we elaborate Cyber Humanism into a design lens for AI-enhanced learning. We articulate three interrelated pillars --- \emph{reflexive competence}, \emph{algorithmic citizenship}, and \emph{dialogic design} --- and introduce \emph{prompt-based learning} as a way of operationalising this lens through natural language interaction with generative AI.

\subsection{Reflexive Competence and Cognitive Sovereignty}

Reflexivity is a cornerstone of both humanistic inquiry and advanced learning. In AI-rich environments, we define \emph{reflexive competence} as the capacity of learners and educators to critically examine how AI systems participate in their cognitive processes and in the construction of knowledge. It extends traditional metacognition by explicitly including reflection on the roles, limits, and affordances of computational agents. From a cyber-humanist standpoint, reflexive competence underpins \emph{cognitive sovereignty}: the ability to retain meaningful control over what one delegates to AI and why, and to recognise when such delegation threatens to erode understanding or mask epistemic uncertainty \cite{CHManifesto}. Designs that foreground comparison, justification, and documentation of AI-assisted work aim to ensure that shortcuts do not come at the cost of diminished epistemic agency.

\subsection{Algorithmic Citizenship in Educational Practice}

While reflexive competence focuses on awareness of human--AI entanglements, \emph{algorithmic citizenship} emphasises the civic and infrastructural dimensions of these entanglements. Rooted in Cyber Humanism, it refers to the rights and responsibilities of individuals and communities vis-à-vis the algorithmic systems that shape their opportunities and obligations. In education, algorithmic citizenship spans classroom practices, course-level norms, and institutional strategies: from understanding how recommender systems or generative feedback tools influence learning trajectories, to participating in the formulation of rules for AI use in assessment, to engaging with debates about procurement, data governance, and compliance with regulations such as the EU AI Act. A cyber-humanist design lens invites educators to treat AI-related rules, settings, and infrastructural choices as legitimate objects of inquiry and co-decision in education.

\subsection{Dialogic Design and Human--AI Collaboration}

The third pillar, \emph{dialogic design}, addresses the interactional dimension of AI-enhanced learning. Dialogic approaches in the Learning Sciences emphasise the co-construction of knowledge through open, multi-voiced dialogue \cite{Wells99}. When AI systems enter this space, they become additional voices that can introduce alternative formulations, examples, or lines of reasoning. From a cyber-humanist perspective, the key question is how to structure such human--AI dialogues so that they expand, rather than constrain, the range of epistemic possibilities. Dialogic design therefore refers to interaction patterns that preserve human agency, foster critical engagement, and make room for contestation, in contrast to designs that encourage uncritical acceptance of AI outputs. Concretely, this may involve asking AI systems for multiple, contrasting solutions and having learners compare and critique them; scripting sequences of prompts that require justification and evidence; or revealing, where feasible, aspects of a model's uncertainty and bias to invite discussion of their implications. In all cases, AI is positioned not as an oracle, but as a fallible interlocutor whose contributions are subject to human scrutiny and reinterpretation.

\subsection{Prompt-Based Learning as a Bridge Between Natural Language and Computational Thinking}

Prompt-based interaction with generative models offers a particularly fertile arena for operationalising these three pillars. We use the term \emph{prompt-based learning} (PBL) to refer to pedagogical designs in which the crafting, sequencing, and analysis of prompts and AI responses become central learning activities, rather than incidental interface operations \cite{PromptPBL}. In this sense, PBL is not merely about ``learning how to prompt'', but about using natural language as a medium for expressing, inspecting, and refining problems and solutions in collaboration with AI.

Natural language interaction with AI can function as a bridge between everyday reasoning and formal computational thinking. When learners describe a problem in natural language to a language model, they are already engaged in decomposition, abstraction, and representation. Through iterative prompting and analysis of AI responses, they can refine these descriptions, identify assumptions, translate narratives into procedures, and, when appropriate, move towards pseudo-code or executable code. The AI system acts as a responsive environment that provides feedback on the clarity and operationalisability of their formulations \cite{DigComp30,AILitFramework}.

Within a cyber-humanist design, PBL is structured to cultivate reflexive competence, algorithmic citizenship, and dialogic design simultaneously. Learners are asked not only to obtain answers, but to document how different prompts lead to different solutions, to explain why certain prompts are more effective or trustworthy, and to reflect on what remains opaque in the AI's behaviour. They interrogate the data assumptions and potential biases behind AI-generated suggestions, discuss who is represented or marginalised in training data, and co-define rules about acceptable uses of AI-generated code or text in their coursework. Educators orchestrate sequences of prompts and counter-prompts that stage the AI as one interlocutor among others (peers, teachers, disciplinary sources), inviting corroboration and challenge. The next section illustrates how these ideas have been translated into higher education practices through courses in social sciences, digital humanities, and engineering, and through a Conversational AI Educator certification within the EPICT ecosystem.

\section{Case Studies in Higher Education
}
\label{sec:casestudies}
This section presents case studies from higher education that operationalise Cyber Humanism through prompt-based learning and professional certification. We first describe the design of a \emph{Conversational AI Educator} certification within the EPICT ecosystem \cite{EPICT}, then three course-level implementations of prompt-based learning across social sciences, digital humanities, and engineering.

\subsection{The Conversational AI Educator and the EPICT Certification}

EPICT (European Pedagogical ICT Licence) is a long-standing certification ecosystem that supports educators in integrating digital technologies into teaching and learning. Building on this tradition, a new \emph{Conversational AI Educator} pathway has been developed to address the specific challenges and opportunities of generative conversational systems in education. It recognises that large language models and related tools are becoming pervasive in learners' practices and in institutional infrastructures \cite{iceri2024}, and that educators require dedicated competences to navigate and shape this landscape.

The Conversational AI Educator profile combines pedagogical design expertise, AI literacy, and a cyber-humanist orientation towards human--AI collaboration. Educators in this role are expected to design prompt-based learning activities that cultivate reflexive competence, to mediate between institutional AI policies and classroom practice, and to support students in developing responsible and critical uses of conversational AI. The EPICT syllabus is structured into progressive levels. At an \emph{Integrator} level, educators incorporate conversational AI into existing courses in a principled way, aligning activities with learning outcomes and assessment criteria and ensuring basic ethical and legal compliance. At an \emph{Expert} level, they design and evaluate more complex scenarios in which AI is used for inquiry, collaboration, and creative tasks, with explicit attention to metacognitive and epistemic goals. At a \emph{Leader} level, educators act as change agents, contributing to institutional strategies, mentoring peers, and coordinating AI-related initiatives. Areas such as foundations of conversational AI, prompt design, AI-enhanced pedagogy, assessment and feedback, ethics and data protection, and community building are mapped onto frameworks including DigCompEdu, DigComp 3.0, and UNESCO's AI competency frameworks for teachers, positioning the certification as an implementation layer that translates high-level descriptors into concrete professional practices.

From a cyber-humanist perspective, the Conversational AI Educator certification makes visible the new forms of expertise required to design and govern AI-rich learning environments and legitimises educators' agency in negotiating how AI is used in their institutions. It also creates a community of practice where issues of reflexive competence, algorithmic citizenship, and dialogic design can be collectively explored.

\subsection{Prompt-Based Learning Across Disciplines}

While certification initiatives target educators' professional development, the everyday enactment of Cyber Humanism takes place in classroom and laboratory activities. This subsection presents three higher education courses that adopt prompt-based learning designs to connect natural language interaction with AI, problem solving, and computational thinking. Although they differ in disciplinary focus, they share a common orientation: treating prompts and AI responses as objects of reflection and negotiation rather than invisible interface details.
\vspace{-8pt}
\paragraph{Social sciences: gamified digital and AI literacy.}

In a social sciences course on digital and AI literacy, generative AI is used both as an object of study and as a tool for inquiry. Students engage in gamified challenges in which they use a conversational AI system to summarise complex texts, identify argumentative structures, or generate alternative framings of social issues. Each challenge is structured around prompt design: students propose prompts, compare outputs, and discuss which prompts better align with criteria such as accuracy, fairness, and rhetorical effectiveness. Meta-tasks require them to document prompting strategies, reflect on how AI responses shape their understanding of source material, and identify hallucinations, biases, or omissions. In this way, prompt-based learning supports reflexive competence and introduces elements of algorithmic citizenship as students negotiate norms for responsible use of AI-generated text in coursework.
\vspace{-8pt}
\paragraph{Digital humanities: natural language as a problem-solving environment.}

In a digital humanities course, prompt-based learning is used to explore unsupervised text analysis and clustering. Students formulate research questions about a corpus (e.g., literary works, policy documents) and describe in natural language how they would group or compare texts to address those questions. They then interact with a conversational AI system to translate these informal descriptions into candidate features, similarity metrics, and clustering approaches, progressively moving towards pseudo-code and code implementations. The iterative dialogue makes explicit how different problem framings lead to different analytical choices and outcomes, and invites critical discussion of the epistemic and bias-related implications of the model's suggestions. Here, natural language functions as a modelling medium, and the human--AI interaction acts as a bridge between conceptual reasoning and computational thinking.
\vspace{-8pt}
\paragraph{Engineering: embedded systems and AI-assisted coding.}

In an engineering course on embedded systems, generative AI and code assistants are integrated into lab activities where students design and implement small-scale hardware--software solutions. Students use AI tools to generate code snippets, explore alternative implementations, and obtain explanations of unfamiliar constructs, but assignments are structured so that AI-generated code is never accepted at face value. Students must specify requirements and constraints in prompts, test and debug the generated code, compare AI-produced solutions with their own or those of peers, and reflect on trade-offs between efficiency, readability, and maintainability. Reflective reports document how AI was used, which prompts were most effective, and where the AI failed to meet requirements. This approach turns AI-assisted coding into an occasion for strengthening understanding of core concepts and for cultivating a sense of responsibility for the final artefacts.

Across these cases, prompt-based learning links everyday language, disciplinary concepts, and computational thinking, while making visible the epistemic moves underpinning AI-mediated work. It provides a concrete arena in which reflexive competence, algorithmic citizenship, and dialogic design can be enacted in higher education practice.

\section{Reclaiming Agency in AI-Rich Educational Ecologies}
\label{sec:discussion}
The cases presented in Section~\ref{sec:casestudies} show how Cyber Humanism can be translated into concrete pedagogical and professional practices. In this section, we discuss how these practices enact the three pillars of reflexive competence, algorithmic citizenship, and dialogic design, and how they contribute to answering the research questions posed in Section~\ref{sec:introduction}. We then highlight key tensions and risks that complicate attempts to reclaim agency in AI-rich learning environments, and outline implications for the Learning Sciences.

\subsection{Enacting Reflexive Competence, Algorithmic Citizenship, and Dialogic Design}

Across the EPICT Conversational AI Educator certification and the prompt-based learning courses, AI is foregrounded not only as an instrument for task completion, but as an object and partner of inquiry. This design choice is central to the cultivation of \emph{reflexive competence}: students document and compare prompting strategies, analyse hallucinations and biases, and articulate criteria for evaluating AI-generated outputs; educators are trained to decide when and why to delegate cognitive tasks to AI and how to render such delegation pedagogically productive. In both cases, human--AI interaction becomes a site where epistemic moves are made visible and open to reflection.

\emph{Algorithmic citizenship} is enacted when learners and teachers participate in negotiating rules for AI use, disclosure, and acceptable levels of assistance, and when educators engage with institutional debates on AI strategies and policies. These practices situate them as stakeholders in the governance of AI in their educational contexts, rather than as passive recipients of top-down regulations. \emph{Dialogic design} operates at the interface between human and AI voices: prompt-based activities are structured to elicit multiple, sometimes conflicting, AI responses and to require comparison, critique, and reframing, preventing models from being treated as unquestioned authorities and positioning them instead as one voice among others in a broader knowledge-building conversation.

Taken together, these elements suggest that Cyber Humanism can indeed reframe the role of AI in education (RQ1), and that specific competencies and frameworks---such as those embodied in the Conversational AI Educator profile and in prompt-based learning designs---can support educators and learners as epistemic and algorithmic citizens in AI-rich environments (RQ2 and RQ3).

\subsection{Tensions, Risks, and Trade-offs}

The case studies also surface tensions that complicate any straightforward narrative of ``reclaiming agency''. A first tension concerns the balance between augmentation and outsourcing of cognitive work: the same tools that can make epistemic processes explicit can also enable \emph{epistemic automation} if learners use them to bypass critical steps of understanding. Careful scaffolding is needed to ensure that prompt-based activities lead to deeper engagement rather than superficial completion.

A second tension relates to workload and professional expectations. Designing cyber-humanist learning activities, with explicit attention to prompts, reflection, and governance, requires time and iterative refinement. In contexts where AI adoption is driven by efficiency or cost-cutting, there is a risk that educators are expected to ``do more with AI'' without adequate support, exacerbating workload and digital stress \cite{EUOSHAAI}. Cyber Humanism presupposes educator agency in shaping AI-rich environments, but this agency depends on institutional recognition and resourcing of the associated design work.

A third tension involves equity and inclusion. Access to AI tools and supportive learning environments is uneven across students and institutions, and the training data of generative models often reflect historical biases and epistemic exclusions. Prompt-based learning can help surface and critique such biases, but it can also normalise them if activities focus mainly on efficiency and performance. Algorithmic citizenship therefore requires attention not only to local classroom norms, but also to the structural conditions and power asymmetries embedded in AI infrastructures.

There are also epistemic risks: the fluency of AI-generated text can subtly reshape students' standards for evidence and argumentation. Dialogic design mitigates this by insisting on corroboration, multi-source comparison, and explicit justification, but such practices must be systematically integrated into tasks and assessment. From a Learning Sciences perspective, this underscores the importance of aligning cyber-humanist designs with robust models of epistemic cognition and disciplinary literacy.

\subsection{Implications for the Learning Sciences}

The experiences discussed here have several implications for the Learning Sciences. First, they indicate that generative AI can be integrated into designs that foreground epistemic agency, rather than undermining it by default. Prompt-based learning, framed within Cyber Humanism, aligns with established concerns about making thinking visible, supporting metacognition, and orchestrating productive dialogue, while adding a focus on AI systems as epistemic artefacts whose behaviour can be interrogated and reshaped.

Second, the Conversational AI Educator certification suggests how new forms of expertise and identity may be formalised and recognised within existing competence frameworks. This opens research questions about how such roles affect teacher agency and wellbeing, how they interact with institutional AI strategies, and how they might be supported through policy and professional development.

Third, the notion of algorithmic citizenship brings civic and governance dimensions to the fore. It invites researchers to examine not only how AI affects individual learning outcomes, but also how learners and educators participate in shaping the rules, norms, and infrastructures of AI use in their institutions and communities, bridging micro-level interactional analyses with meso- and macro-level studies of policy, labour, and infrastructure.

Finally, studying cyber-humanist practices will likely require mixed-methods approaches that combine fine-grained analysis of human--AI interaction (e.g., prompt logs, conversation transcripts) with ethnographic accounts of institutional decision-making and longitudinal studies of competence development. By engaging with these complexities, the Learning Sciences can contribute to ensuring that the integration of AI advances, rather than diminishes, democratic and human-centred aims in education.

\section{Conclusion}
\label{sec:conclusion}
This paper has argued that the widespread adoption of generative AI in education calls for a shift from viewing AI as an external disruptive force or neutral tool towards understanding it as part of the cognitive and institutional infrastructures of learning. Building on the paradigm of Cyber Humanism, we proposed a design lens that foregrounds learners and educators not merely as users of AI, but as epistemic agents and algorithmic citizens who can interrogate, shape, and co-govern AI-rich learning environments through \emph{reflexive competence}, \emph{algorithmic citizenship}, and \emph{dialogic design} (RQ1).

In addressing RQ2, we positioned this lens alongside major digital and AI competence frameworks, including DigComp 3.0, DigCompEdu, UNESCO's AI competency frameworks, and the OECD/EC AI Literacy framework. While these frameworks recognise AI literacy as multi-dimensional, our analysis highlighted gaps concerning infrastructural participation, the everyday enactment of algorithmic citizenship, and the role of natural language as a modelling medium in human--AI interaction. To respond to RQ3, we presented case studies from higher education that operationalise Cyber Humanism through prompt-based learning and a Conversational AI Educator certification within the EPICT ecosystem. These examples showed how natural language interaction with GenAI can serve as a bridge between problem description, problem solving, and computational thinking, and how new professional profiles can support educators in designing and governing AI-rich environments.

The cases also exposed tensions around cognitive outsourcing, workload and wellbeing, equity of access, and the reproduction of biases through AI models and platforms, indicating that reclaiming agency in AI-rich education is an ongoing negotiation rather than a one-off design choice. For the Learning Sciences community, the cyber-humanist perspective opens a research agenda that includes tracing how epistemic agency is negotiated in human-AI dialogues, examining how roles such as the Conversational AI Educator evolve within institutional settings, and studying how learners and educators participate in the multi-level governance of AI systems. Within the broader WAILS agenda, Cyber Humanism in Education offers a way to think both \emph{with} and \emph{for} the Learning Sciences: drawing on established theories of epistemic cognition and dialogic learning to shape human-AI collaboration, and recognising AI as a force that reshapes the very environments in which learning and educational research take place.

%
%
%
%

\end{document}